\pdfoutput=1
\documentclass[11pt]{article}
\usepackage[T1]{fontenc}
\usepackage[utf8]{inputenc}
\usepackage[a4paper,margin=1in]{geometry}
\usepackage{times}
\usepackage{microtype}
\usepackage{natbib}
\usepackage{float}
\usepackage{amsmath,amsfonts,bm}

\def\eqref#1{equation~\ref{#1}}
\def\1{\bm{1}}

\DeclareMathAlphabet{\mathsfit}{\encodingdefault}{\sfdefault}{m}{sl}
\SetMathAlphabet{\mathsfit}{bold}{\encodingdefault}{\sfdefault}{bx}{n}

\usepackage[hypertexnames=false]{hyperref}
\usepackage{url}
\usepackage{placeins}
\usepackage{needspace}
\hypersetup{hidelinks,pdftitle={Bridging the Evidence-to-Execution Gap: A Reflective Agent for Multi-Objective Peptide Design},pdfauthor={Haosen Zhang; Yang Yang}}
\usepackage{booktabs,graphicx,amsmath,amssymb,bm,array}
\usepackage{etoolbox,tabularx,siunitx}
\usepackage[table]{xcolor}
\usepackage{caption}
\definecolor{tableShade}{gray}{0.95}
\definecolor{tableGroup}{gray}{0.32}
\newcommand{\tablehead}[1]{{\footnotesize\bfseries #1}}
\newcommand{\tablegroup}[1]{{\footnotesize\bfseries\color{tableGroup} #1}}
\AtBeginEnvironment{table}{\rmfamily\small\setlength{\tabcolsep}{5pt}}
\usepackage{algorithm,algpseudocode}

\newcommand{\rwh}[1]{\par\noindent\textbf{#1}\ }
\graphicspath{{figures/}}
\title{Bridging the Evidence-to-Execution Gap:\\A Reflective Agent for Multi-Objective Peptide Design}
\author{%
Haosen Zhang$^{1}$ \qquad Yang Yang$^{2,*}$\\
\small $^{1}$School of Agriculture and Biology, Shanghai Jiao Tong University\\
\small $^{2}$School of Computer Science, Shanghai Jiao Tong University\\
\small $^{*}$Corresponding author
}
\date{}
\begin{document}
\pagestyle{plain}
\maketitle

\begin{abstract}
Large language models can reason over scientific literature, but textual plans do not reliably produce biological sequences with the intended properties. Sequence generators capture sequence patterns, yet they do not interpret evidence, choose interventions, or reflect on outcomes. This separation creates an evidence-to-execution gap between scientific reasoning and sequence-level control. We introduce EASER, a reflective agent that connects the two through a learned property interface. The interface consists of property-specific low-rank directions trained offline and fixed during search; an LLM controller (the controller) composes them to steer a frozen diffusion generator. From retrieved evidence, sequence context, and prior outcomes, it proposes intervention hypotheses that specify anchors, editable positions, and coefficients. Probe-and-Steer tests these hypotheses and allocates further samples from their joint predicted responses; reflection updates the agent's memory. On computational antimicrobial peptide design, EASER balances activity, non-hemolysis, and non-toxicity. Explicit hypothesis formulation yields larger multi-objective gains than direct action generation, and ablations support evidence retrieval, episodic history, reflection, and Probe-and-Steer. EASER achieves the highest mean hypervolume and lowest mean IGD+ among the evaluated search policies and generators on screened candidates. These results show that a learned property interface can make literature-grounded reasoning executable at the sequence level.
\end{abstract}

\section{Introduction}
Scientific literature offers interventional principles for biological sequence design, specifying changes and the conditions under which they may help. Residue substitutions, for example, can alter both antibacterial activity and hemolysis \citep{chenhydrophobicity2007}. Applying this knowledge requires an interface for making executable changes to a particular sequence.

Scientific evidence constrains which sequence changes are plausible, but it does not specify how to execute them in a generative model. In antimicrobial peptide (AMP) design, activity must be balanced against hemolysis and toxicity, and the same hydrophobic substitution can improve one property while degrading another \citep{chenhydrophobicity2007}. A useful system must connect a retrieved principle to a concrete edit, observe the joint response, and revise the next edit. This is the evidence-to-execution gap we address.

Existing approaches connect language and generation through paired text--sequence learning, reward-based optimization, or tool-using agents. ProteinDT aligns text and protein representations, while ProtDAT conditions generation on protein descriptions \citep{proteindt2025,protdat2025}. Reward and preference methods specify desired outcomes \citep{drakes2025,proteindpo2026}, while ProtoCycle combines reflective planning with tool feedback \citep{ge2026protocycle}. We study an interface through which an agent uses retrieved evidence and observed responses to select property-specific directions, sequence anchors, and editable positions.

\begin{figure}[!t]
\centering
\includegraphics[width=\linewidth]{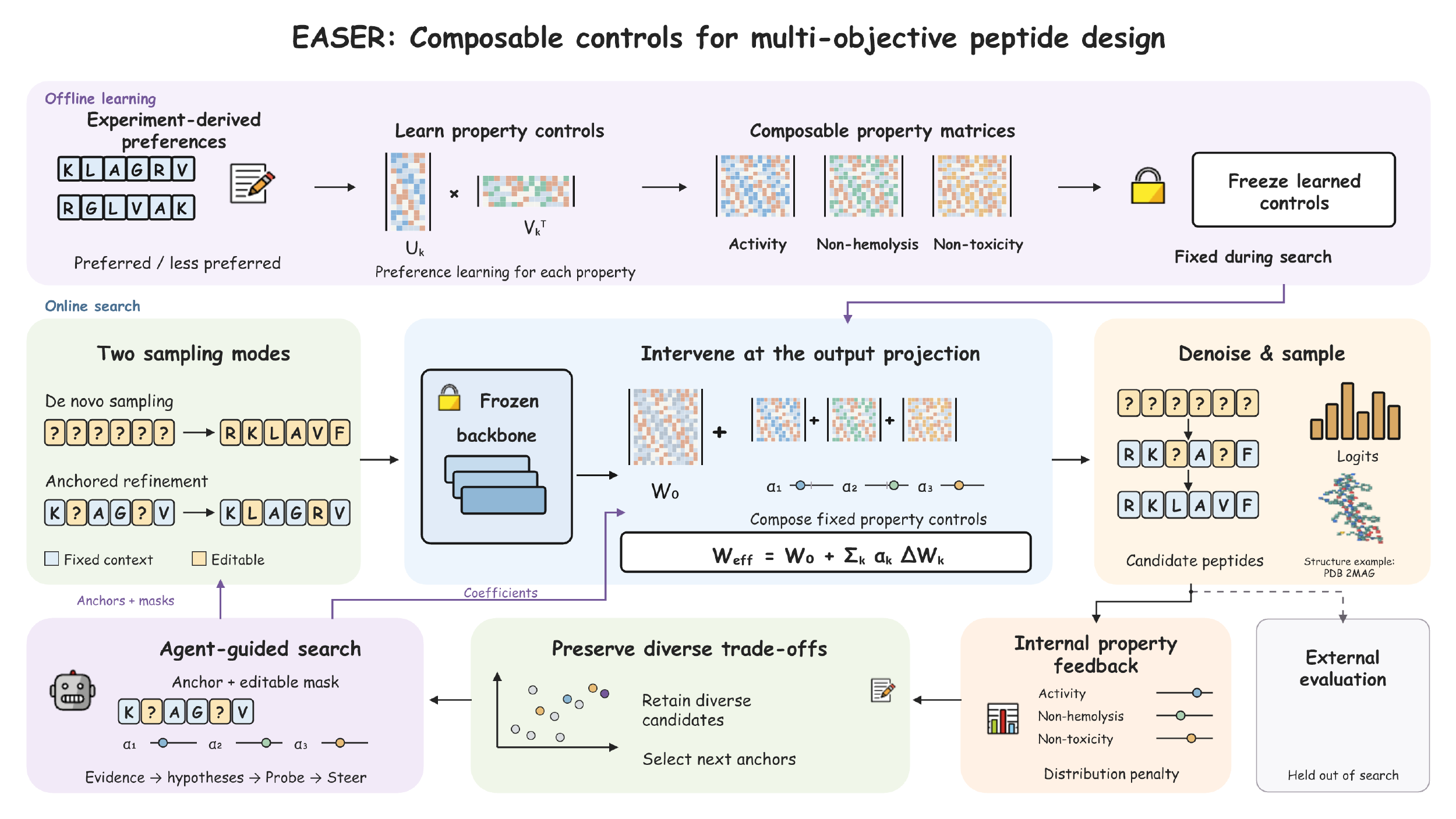}
\caption{\textbf{EASER overview.} Offline-trained matrices steer a frozen generator during sampling and refinement. Evidence and internal predictor feedback guide agent decisions; external predictors score candidates after search. The diagram uses schematic matrices, sequences, and plots; the molecular example is \href{https://www.rcsb.org/structure/2MAG}{PDB 2MAG}.}
\label{fig:framework}
\end{figure}

We introduce EASER (Evidence-Aware Sequence Engineering with Reflection), a reflective agent with composable directions over a frozen peptide diffusion generator (Figure~\ref{fig:framework}). We learn property-specific low-rank matrices offline at the output projection. The LLM controller combines them through bounded coefficients while selecting anchors and editable positions. The matrices define the available operations; evidence guides the agent's choice of when and where to apply them.

Because directions interact, the controller treats interventions as testable hypotheses. Probe-and-Steer tests small batches and allocates further samples from their joint predicted responses. Reflection links actions to outcomes for subsequent decisions. The controller checks literature-derived expectations against computational feedback while keeping the generator and matrices fixed.

We test whether interventions chosen from evidence improve AMP search using matched-state comparisons, component ablations, and analyses of coupled property responses. With a budget of 2,560 candidates per method, EASER achieves the highest mean screened hypervolume and lowest mean IGD+ among the evaluated policies and generators.

\rwh{Contributions.}
We provide (1) an executable interface of composable low-rank property directions; (2) an evidence-grounded reflective agent that tests interventions through Probe-and-Steer; and (3) empirical validation of improved AMP search and intervention decisions, supported by component ablations and analysis of coupled responses.

\FloatBarrier
\section{Related Work}
\rwh{Generative peptide models.}
AMPGAN v2 and AMPGPT2 use adversarial and autoregressive generation \citep{ampgan2021,ampgpt22026}, while ProT-Diff and AMPGen use diffusion \citep{protdiff2024,ampgen2025}. AMP-Designer combines prompt tuning and reinforcement learning \citep{ampdesigner2024}; HMAMP and MOFormer address multiple objectives through hypervolume optimization and Pareto ranking \citep{hmamp2025,moformer2025}. EASER focuses on the agent's decision process: using domain evidence and search history to select interventions and revising decisions from observed responses.

\rwh{Controllable generation.}
Preference optimization learns from paired outputs; DRAKES combines rewards with divergence regularization, and ProteinDPO uses experimental stability preferences \citep{dpo2023,drakes2025,proteindpo2026}. TR2-D2 trains on tree-selected trajectories, while token-aligned preference optimization learns property-conditioned preferences \citep{tr2d22025,tokenaligned2025}. Classifier and classifier-free guidance steer diffusion sampling; PepTune and MP2D incorporate multi-objective tree search \citep{classifierguidance2021,classifierfree2022,peptune2025,mp2d2026}. LoRA, model soups, and Rewarded Soups support compact updates and combinations of adapted models \citep{lora2022,modelsoups2022,rewardedsoups2023}. These controls express desired outcomes through rewards, labels, conditioning, or predictor-guided search. EASER connects residue- and property-level evidence to decisions about anchors, editable positions, and coefficients for property-trained matrices.

\rwh{Scientific agents.}
ReAct interleaves reasoning with actions, while Reflexion incorporates textual feedback into subsequent decisions \citep{react2023,reflexion2023}. In scientific applications, ChemCrow and Coscientist connect language models to chemistry tools and experimental execution, and SciToolAgent uses a tool knowledge graph to support workflow selection \citep{chemcrow2024,coscientist2023,scitoolagent2025}. ProtAgents, OSDA Agent, and the Virtual Lab coordinate domain expertise and tools for scientific tasks \citep{protagents2024,osdaagent2025,virtuallab2025}. For peptide design, PepCraft, introduced with AMPGAN v3, coordinates generation and verification, while Pepti-Agent connects generation, prediction, and mutation through feedback \citep{ampganv32026,peptiagent2026}. ProtoCycle couples reflective planning with scaffold retrieval, functional-site design, and evaluation, and trains its planner through supervised fine-tuning and online reinforcement learning \citep{ge2026protocycle}. These systems connect planning, tools, and feedback in scientific workflows. In multi-objective sequence design, evidence can support competing modifications to the same candidate. EASER addresses this decision by testing hypotheses through learned property directions and revising subsequent interventions from their joint responses.

MAC-AMP connects multi-agent assessment to generator adaptation through dynamically designed rewards and successive PPO updates \citep{macamp2026}. Reward selection uses Pareto comparisons, but each PPO stage optimizes a scalar objective that need not preserve diverse activity--safety trade-offs. EASER tests evidence-informed interventions on a frozen generator and uses joint property responses to guide sampling.

\section{Method}
\label{sec:method}
EASER searches for favorable predicted activity--safety trade-offs under a fixed generation budget. We call the property-specific low-rank directions and their coefficients the property interface. An intervention is one executable choice of anchors, editable positions, coefficients, and strength. Throughout, non-hemolysis and non-toxicity denote the evaluation objectives, while low-hemolysis and low-toxicity denote the corresponding trained directions. The controller selects interventions; EASER denotes the full agent, including retrieval, evidence aggregators, memory, and the controller. The controller uses evidence and prior outcomes to choose interventions, then uses observed responses to guide further sampling (Figure~\ref{fig:framework}).

\subsection{Low-rank output interventions}
Low-rank updates at the output projection alter amino-acid sampling preferences while the pretrained transformer supplies contextual features. Weighted combinations apply several property directions in one generation call.

Our AMP-adapted Diffusion Protein Language Model (DPLM, 650M parameters) \citep{dplm2024} undergoes sequential LoRA adaptation on short peptides and an AMP-oriented corpus (Appendix~\ref{app:backbone-adaptation}). It remains frozen during direction training and search; Appendix~\ref{app:data-sources} describes the data. For contextual state $h_{i,t}\in\mathbb R^d$ at position $i$ and denoising step $t$, the base logits are $W_0h_{i,t}+b$. For each property $k\in\{a,h,t\}$, we learn
\begin{equation}
\Delta W_k=U_kV_k^\top,\qquad U_k\in\mathbb R^{v\times r},\quad V_k\in\mathbb R^{d\times r}.
\label{eq:factor}
\end{equation}
With $v=33$, $d=1280$, and $r=16$, the three factor pairs contain $3r(v+d)=63{,}024$ trainable parameters. Once learned, these matrices remain fixed while the controller changes their coefficients during search.

Bounded multipliers $\alpha_k$ combine fixed, norm-calibrated matrices (Appendix~\ref{app:training}):
\begin{equation}
W_{\mathrm{eff}}=W_0+\sum_k\alpha_k\Delta W_k,
\qquad \ell_{i,t}=W_{\mathrm{eff}}h_{i,t}+b.
\label{eq:intervention}
\end{equation}
Zero coefficients recover the base projection; calibration is fixed across comparisons. Anchored refinement masks and denoises selected residues while preserving the rest; de novo sampling masks every position. The contextual intervention $\Delta W_kh_{i,t}$ varies by position and denoising step. Logits are linear in coefficients at fixed hidden states. Generated properties also depend on normalization, denoising, and sampling, so we evaluate their responses empirically.

\begin{figure}[!htbp]
\centering
\includegraphics[width=\linewidth]{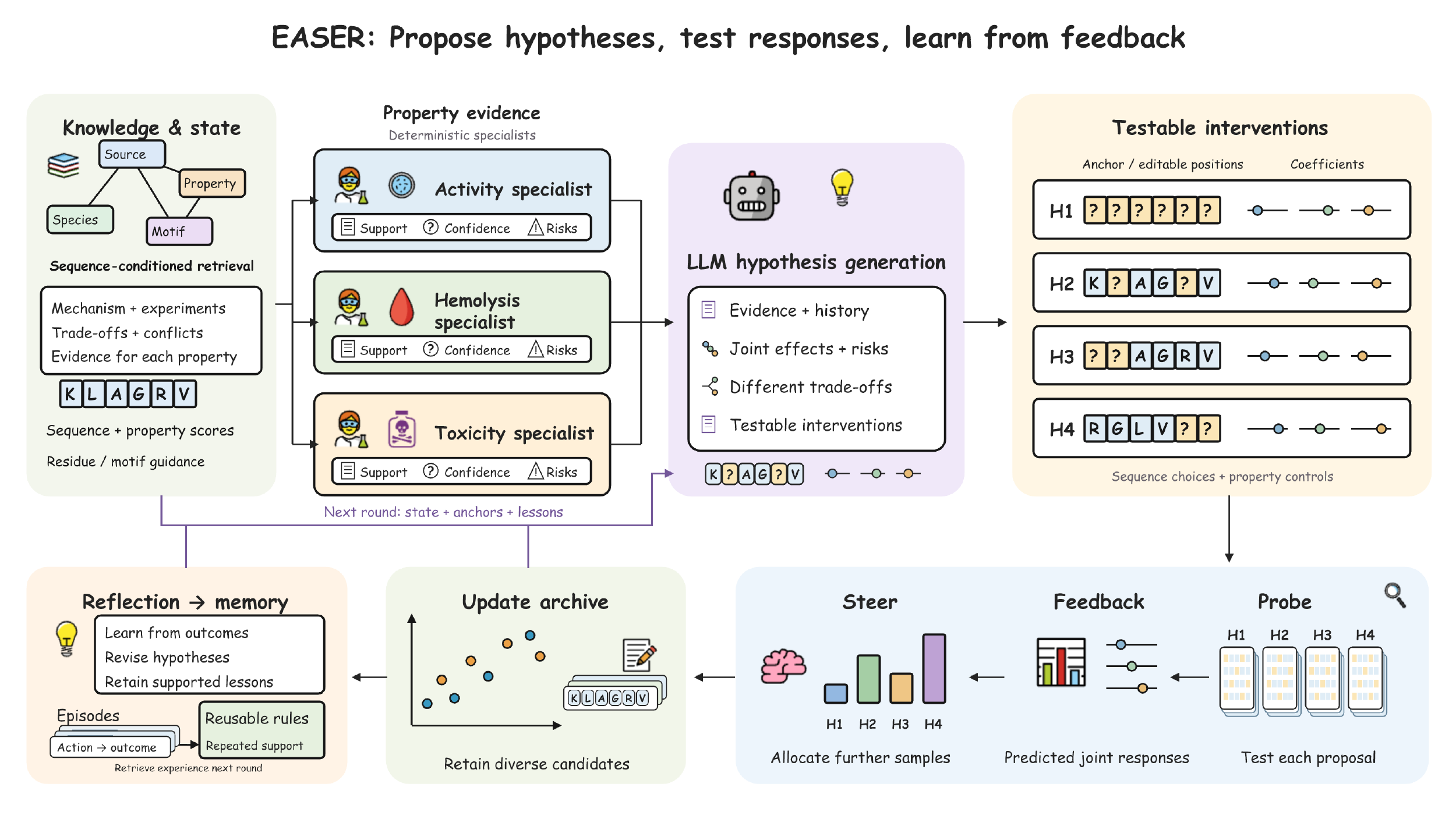}
\caption{\textbf{Evidence-informed agent loop.} Evidence aggregators organize evidence; the controller proposes interventions, Probe tests responses, and Steer allocates samples. Outcomes update the archive, reflection, and memory. Sequences, directions, and plots are schematic.}
\label{fig:agent-framework}
\end{figure}

\Needspace{7\baselineskip}
\subsection{Evidence-informed intervention hypotheses}
Sequence-conditioned retrieval supplies organism, residue, and motif evidence to three property-specific evidence aggregators (Figure~\ref{fig:agent-framework}). The aggregators combine supporting and adverse evidence deterministically; the controller generates joint proposals, allocates samples, and reflects on outcomes.

The controller combines candidate sequences, predicted properties, physicochemical summaries, aggregator recommendations, and previous outcomes. It uses these inputs to form intervention hypotheses. Each hypothesis specifies anchor sequences $P$, editable positions $M$, coefficients $\bm c$, and shared strength $\eta$, together with expected effects and risks. For example, modifying a hydrophobic region may reduce hemolysis while risking a loss of activity. An empty anchor set specifies controlled de novo sampling. Appendix~\ref{app:knowledge-architecture} describes the evidence resource.

\rwh{Round summary.}
Each round takes the archive, the previous reflection, and 32 fresh de novo candidates as input. (1) Build the round context. (2) Retrieve evidence and form four hypotheses $(P_j,M_j,\bm c_j,\eta_j)$. (3) Probe 16 candidates per hypothesis and score them with the internal evaluator. (4) Allocate 160 Steer samples across hypotheses from their joint responses. (5) Update the archive, summarize outcomes, and update memory for the next round.

\begin{quote}
\small
\noindent\rule{0.96\linewidth}{0.4pt}\par
\noindent\textbf{Worked example.} The controller targeted a C-terminal hydrophobic region with coefficients $(0.65,0.80,0.90)$. Probe tested four hypotheses with 16 candidates each; Steer then assigned 0, 0, 40, and 120 samples to them. The selected offspring replaced nine residues and raised predicted non-hemolysis from 0.4792 to 0.6811 before penalty while activity stayed near 0.65.
\par\noindent\rule{0.96\linewidth}{0.4pt}
\end{quote}

Appendix~\ref{app:agents} gives the full procedure; Appendix~\ref{app:search-case} traces a recorded run.

\subsection{Learning the property directions}
We learn each direction offline from preferred and dispreferred peptide pairs matched by coarse physicochemical descriptors. Activity preferences use MIC, the low-hemolysis direction uses HC50, and the low-toxicity direction uses curated labels. We train the low-rank factors with a reference-relative masked preference objective using shared diffusion corruption, replay, and norm regularization. We select checkpoints by validation performance and calibrate their norms before search. Appendix~\ref{app:training} details the objective, preference pairs, and training settings.

\FloatBarrier
\section{Experimental Setup}
\label{sec:setup}
\rwh{Task and comparisons.}
We compared EASER with five generators (AMPGAN v2, AMPGPT2, AMPGen, AMP-Designer, and AMP-adapted DPLM) on predicted AMP activity, non-hemolysis, and non-toxicity. To examine intervention selection, we compared EASER with fixed coefficients, random actions, LinUCB, and NSGA-II using the same frozen generator and matrices. LinUCB and NSGA-II selected anchors, masks, coefficients, and strength within the shared action space. Component ablations and matched-information comparisons examined the roles of evidence and feedback; random-matrix controls tested the learned interventions.

\rwh{Evaluation protocol.}
Main comparisons used a budget of 2,560 candidates per method, with lengths of 15--35 residues and ten rounds of 256 candidates. Component ablations and full-search matrix replacement were paired with EASER. Internal predictors supplied feedback; external predictors and the evaluation front were withheld from decisions. Common normalization and physicochemical penalties applied to all methods, with screening requiring all three adjusted utilities above 0.5 (Appendices~\ref{app:evaluation} and~\ref{app:statistical-protocol}).

\rwh{Metrics.}
Hypervolume (HV) and modified inverted generational distance (IGD+) \citep{igdplus2015} measured joint performance on screened candidates using normalized scores before penalty. HV used reference $(0.5,0.5,0.5)$; IGD+ used a fixed empirical front (Appendix~\ref{app:statistical-protocol}). Property-specific top-ten means used separate selected sets. We report equal-weight run means and sample SD; sensitivity analyses reuse the same candidates.

\section{Results}

\subsection{Multi-objective peptide design}
EASER improved joint activity--safety performance under the common evaluation protocol (Table~\ref{tab:external-generators}). It achieved the highest mean HV, 0.0450 versus 0.0301--0.0362 for the baseline generators, and the lowest mean IGD+.

\begin{table}[!htbp]
\centering\small
\caption{\textbf{External-generator property scores.} Mean $\pm$ sample SD of per-run top-ten means, after screening and before penalty. Each property selects its own sequences.}
\label{tab:external-properties-main}
\begin{tabularx}{\linewidth}{>{\raggedright\arraybackslash}XS[table-format=1.3(3),table-column-width=0.23\linewidth]S[table-format=1.3(3),table-column-width=0.23\linewidth]S[table-format=1.3(3),table-column-width=0.23\linewidth]}
\toprule
\multicolumn{1}{l}{\tablehead{Method}} & \multicolumn{1}{c}{\tablehead{\shortstack{Activity\\score $\uparrow$}}} & \multicolumn{1}{c}{\tablehead{\shortstack{Non-hemolysis\\score $\uparrow$}}} & \multicolumn{1}{c}{\tablehead{\shortstack{Non-toxicity\\score $\uparrow$}}}\\
\midrule
AMPGAN v2 & 0.843 +- 0.019 & 0.704 +- 0.008 & 0.963 +- 0.001\\
AMPGPT2 & 0.908 +- 0.012 & 0.688 +- 0.014 & 0.958 +- 0.001\\
AMPGen & 0.809 +- 0.049 & 0.668 +- 0.016 & 0.962 +- 0.000\\
AMP-Designer & 0.884 +- 0.017 & 0.705 +- 0.014 & 0.963 +- 0.001\\
AMP-adapted DPLM & 0.856 +- 0.029 & 0.689 +- 0.016 & 0.962 +- 0.001\\
\addlinespace[2pt]
\rowcolor{tableShade}
\textbf{EASER} & 0.901 +- 0.067 & 0.731 +- 0.024 & 0.963 +- 0.003\\
\bottomrule
\end{tabularx}
\end{table}

\begin{table}[!htbp]
\centering\small
\caption{\textbf{External-generator frontier quality.} Passed counts candidates meeting all three adjusted utility thresholds. HV/IGD+ report mean $\pm$ sample SD on screened, unpenalized scores with the fixed reference front.}
\label{tab:external-generators}
\begin{tabularx}{\linewidth}{>{\raggedright\arraybackslash}XS[table-format=4,table-column-width=0.09\linewidth]S[table-format=1.4(4),table-column-width=0.215\linewidth]S[table-format=1.4(4),table-column-width=0.215\linewidth]}
\toprule
\multicolumn{1}{l}{\tablehead{Method}} & \multicolumn{1}{c}{\tablehead{Passed}} & \multicolumn{1}{c}{\tablehead{HV $\uparrow$}} & \multicolumn{1}{c}{\tablehead{IGD+ $\downarrow$}}\\
\midrule
AMPGAN v2 & 1100 & 0.0362 +- 0.0044 & 0.0543 +- 0.0182\\
AMPGPT2 & 1717 & 0.0330 +- 0.0026 & 0.0603 +- 0.0090\\
AMPGen & 395 & 0.0301 +- 0.0065 & 0.0739 +- 0.0307\\
AMP-Designer & 1409 & 0.0356 +- 0.0041 & 0.0519 +- 0.0106\\
AMP-adapted DPLM & 560 & 0.0350 +- 0.0042 & 0.0518 +- 0.0131\\
\addlinespace[2pt]
\rowcolor{tableShade}
\textbf{EASER} & 1947 & 0.0450 +- 0.0086 & 0.0277 +- 0.0177\\
\bottomrule
\end{tabularx}
\end{table}

EASER led the joint frontier metrics, while property-specific rankings varied across methods (Table~\ref{tab:external-properties-main}). EASER had the highest non-hemolysis top-ten mean, whereas AMPGPT2 led activity and several generators had similar non-toxicity means at the reported precision. Each property ranking selects its own candidates, whereas HV and IGD+ measure the joint property combinations attained by the generated sets. Appendix~\ref{app:external-physchem} reports their physicochemical distributions.

\subsection{Agent decisions shape the search frontier}
\begin{figure}[!htbp]
\centering
\includegraphics[width=\linewidth]{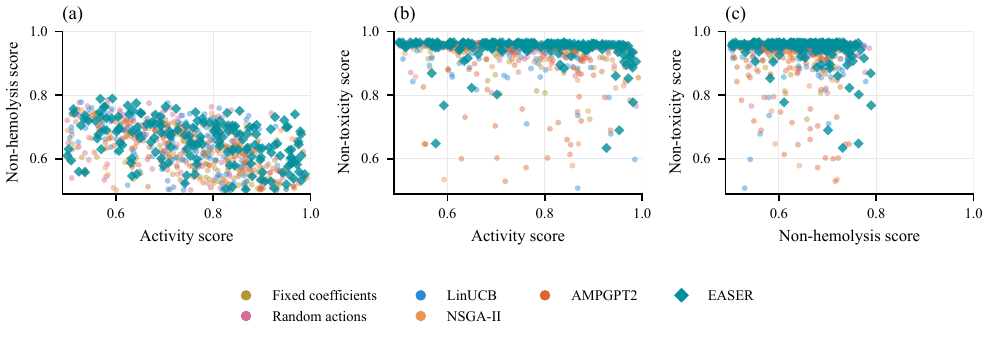}
\caption{\textbf{Pareto-front projections.} Each panel projects the screened three-objective front onto a different pair of objectives (activity, non-hemolysis, and non-toxicity). Points are individual screened candidates; coordinates are normalized scores before penalty. Each method contributes its run-level fronts.}
\label{fig:pareto}
\end{figure}

\rwh{Choosing interventions on a shared generator.}
With the generator and matrices held fixed, EASER achieved the highest mean HV and lowest mean IGD+ (Table~\ref{tab:main}; Figure~\ref{fig:pareto}). Mean HV was 0.0450, approximately 15.7\% above random actions (0.0389), versus 0.0373 for LinUCB and 0.0344 for NSGA-II. The trade-off sets varied with how each policy selected interventions and used the available information, even with the same learned directions (Appendices~\ref{app:search-progress} and~\ref{app:coverage}).

\begin{table}[t]
\centering\small
\caption{\textbf{Search policies on a shared generator and learned directions.} Mean $\pm$ sample SD; each method contributes 2,560 candidates. Metrics use screened, unpenalized scores and the fixed reference (Appendix~\ref{app:statistical-protocol}).}
\label{tab:main}
\begin{tabularx}{\linewidth}{>{\raggedright\arraybackslash}XS[table-format=1.4(4),table-column-width=0.28\linewidth]S[table-format=1.4(4),table-column-width=0.28\linewidth]}
\toprule
\multicolumn{1}{l}{\tablehead{Method}} & \multicolumn{1}{c}{\tablehead{Exact HV $\uparrow$}} & \multicolumn{1}{c}{\tablehead{IGD+ $\downarrow$}}\\
\midrule
Fixed coefficients & 0.0282 +- 0.0076 & 0.0781 +- 0.0329\\
Random actions & 0.0389 +- 0.0068 & 0.0414 +- 0.0162\\
LinUCB & 0.0373 +- 0.0087 & 0.0540 +- 0.0260\\
NSGA-II & 0.0344 +- 0.0069 & 0.0588 +- 0.0240\\
\rowcolor{tableShade}
\textbf{EASER} & 0.0450 +- 0.0086 & 0.0277 +- 0.0177\\
\bottomrule
\end{tabularx}
\end{table}

\rwh{Formulating hypotheses from the same evidence.}
We compared explicit hypothesis formulation with direct action generation at 30 saved decision states. Both protocols used the same controller, sequences, evidence, and history, and proposed four actions with matched sampling schedules and allocations. Explicit formulation additionally stated expected effects, supporting evidence, and risks before specifying the actions.

Explicit formulation increased mean HV gain over each state's existing candidates from 0.001510 to 0.002447, with screened yield rising from 162 to 261 (Table~\ref{tab:grounded-reasoning}). The matched-input comparison favored stating testable expectations before selecting interventions (Appendix~\ref{app:grounded-reasoning}).

\begin{table}[!htbp]
\centering\small
\caption{\textbf{Hypothesis formulation with matched information.} Mean HV gain over pre-existing candidates at 30 states; screened counts sum across 1,920 draws per protocol.}
\label{tab:grounded-reasoning}
\begin{tabularx}{\linewidth}{>{\raggedright\arraybackslash}XS[table-format=1.6,table-column-width=0.23\linewidth]S[table-format=3,table-column-width=0.23\linewidth]}
\toprule
\multicolumn{1}{l}{\tablehead{Policy}} & \multicolumn{1}{c}{\tablehead{Mean $\Delta\mathrm{HV}$ $\uparrow$}} & \multicolumn{1}{c}{\tablehead{\shortstack{Screened\\candidates $\uparrow$}}}\\
\midrule
Direct action generation & 0.001510 & 162\\
\rowcolor{tableShade}
\textbf{Explicit hypothesis formulation} & \bfseries 0.002447 & \bfseries 261\\
\bottomrule
\end{tabularx}
\end{table}

\rwh{Using feedback to revise subsequent decisions.}
Removing reflection feedback reduced mean HV by 14.1\% and screened yield fell to 1,025 (Table~\ref{tab:targeted-feedback}). Removing action and outcome details from round-history inputs, while retaining reflections and retrieved experience, reduced HV by 15.2\% and yield fell to 574 (Appendix~\ref{app:targeted-controls}).

\begin{table}[!htbp]
\centering\small
\caption{\textbf{Targeted LLM-input ablations.} Changes relative to EASER. Passed is pooled across the evaluated candidates per condition. Appendix~\ref{app:targeted-controls} defines the removed inputs.}
\label{tab:targeted-feedback}
\begin{tabularx}{\linewidth}{>{\raggedright\arraybackslash}XS[table-format=+1.4,table-column-width=0.215\linewidth]S[table-format=+1.4,table-column-width=0.215\linewidth]S[table-format=4,table-column-width=0.07\linewidth]}
\toprule
\multicolumn{1}{l}{\tablehead{Condition}} & \multicolumn{1}{c}{\tablehead{$\Delta$HV $\uparrow$}} & \multicolumn{1}{c}{\tablehead{$\Delta$IGD+ ($+$ worse)}} & \multicolumn{1}{c}{\tablehead{Passed}}\\
\midrule
Without reflection feedback & -0.0065 & +0.0165 & 1025\\
Without round-history details & -0.0070 & +0.0146 & 574\\
\bottomrule
\end{tabularx}
\end{table}

\rwh{Testing interventions and using evidence.}
Removing Probe-and-Steer's multi-proposal evaluation and adaptive allocation increased screened yield to 1,089 and reduced mean HV by 33.2\% (Table~\ref{tab:ablation}). Removing retrieval reduced HV by 18.2\% with similar screened yield (867). Jointly removing memory and cross-round anchor reuse reduced HV by 20.2\% (Appendix~\ref{app:targeted-controls}).

\begin{table}[t]
\centering\small
\caption{\textbf{Component ablations and matrix replacement.} Changes relative to EASER. Metrics use screened, unpenalized scores and the common reference; Passed is pooled across the evaluated candidates.}
\label{tab:ablation}
\begin{tabularx}{\linewidth}{>{\raggedright\arraybackslash}XS[table-format=+1.4,table-column-width=0.215\linewidth]S[table-format=+1.4,table-column-width=0.215\linewidth]S[table-format=4,table-column-width=0.07\linewidth]}
\toprule
\multicolumn{1}{l}{\tablehead{Method}} & \multicolumn{1}{c}{\tablehead{$\Delta$HV $\uparrow$}} & \multicolumn{1}{c}{\tablehead{$\Delta$IGD+ ($+$ worse)}} & \multicolumn{1}{c}{\tablehead{Passed}}\\
\midrule
Without memory and anchor reuse & -0.0089 & +0.0248 & 524\\
Without retrieval & -0.0080 & +0.0234 & 867\\
Without Probe-and-Steer & -0.0146 & +0.0382 & 1089\\
All matrices randomized & -0.0094 & +0.0189 & 1076\\
\bottomrule
\end{tabularx}
\end{table}

\subsection{Learned directions produce coupled property responses}
\rwh{Property-specific response.}
Each learned direction raised its target score under positive control in de novo sampling, and all three exceeded their spectrum-matched random controls. Target means increased monotonically across nonnegative coefficients. Appendix~\ref{app:component-diagnostics} reports the full dose grid.

\rwh{Combining directions expands the observed trade-offs.}
Joint settings reached property combinations unavailable in single-direction scans. The learned directions produced higher pooled screened HV than random sets, and many nonzero settings improved each adjusted mean. The coefficient settings that maximized each property mean after KDE adjustment (Gaussian kernel density estimate) differed across objectives. Appendix~\ref{app:full-position-grid} reports the grid, and Appendix~\ref{app:matrix} examines the underlying residue preferences.

\rwh{Direction orientation matters within adaptive search.}
Replacing the learned directions with spectrum- and norm-matched random directions reduced mean search HV and increased screened yield (Table~\ref{tab:ablation}). The controller adapted its actions using feedback from each fixed random set (Appendix~\ref{app:full-search-random}).

\section{Discussion and Conclusion}
EASER connects evidence-grounded decisions to sequence generation through learned property directions. Learned matrices produced larger property responses than spectrum-matched random controls, while hypothesis formulation improved intervention outcomes under matched information. EASER learns reusable sequence operations offline, then uses evidence and feedback to choose interventions during search.

The coefficient grid revealed property-dependent coefficient optima and both favorable and adverse trade-offs. Probe-and-Steer observes these joint responses before allocating further samples. Reflection links the responses to executed actions so later proposals can use this experience.

\rwh{Evaluation scope.}
The experiments evaluated predicted AMP activity--safety trade-offs. AMP design used the complete agent; the matched-state analysis checked cited observations against supplied values and compared proposed masks with their rationales.

\clearpage
\begingroup
\small
\setlength{\bibsep}{3pt}\urlstyle{same}
\bibliography{references}
\bibliographystyle{abbrvnat}
\endgroup
\clearpage
\appendix
\small
\setlength{\textfloatsep}{10pt plus 2pt minus 2pt}
\setlength{\floatsep}{8pt plus 2pt minus 2pt}
\setlength{\intextsep}{8pt plus 2pt minus 2pt}
\setlength{\abovecaptionskip}{5pt}
\setlength{\belowcaptionskip}{2pt}
\setlength{\abovedisplayskip}{7pt plus 2pt minus 2pt}
\setlength{\belowdisplayskip}{7pt plus 2pt minus 2pt}
\makeatletter
\renewcommand\section{\@startsection{section}{1}{\z@}{-2.0ex plus -.4ex minus -.2ex}{.9ex plus .2ex}{\normalfont\large\bfseries}}
\renewcommand\subsection{\@startsection{subsection}{2}{\z@}{-1.4ex plus -.3ex minus -.1ex}{.6ex plus .1ex}{\normalfont\normalsize\bfseries}}
\setlength{\@fptop}{0pt}
\setlength{\@fpsep}{10pt}
\setlength{\@fpbot}{0pt plus 1fil}
\makeatother
\renewcommand{\topfraction}{0.95}
\renewcommand{\bottomfraction}{0.95}
\renewcommand{\textfraction}{0.05}
\renewcommand{\floatpagefraction}{0.80}
\numberwithin{equation}{section}
\renewcommand{\theequation}{\thesection\arabic{equation}}
\raggedbottom
\section*{Appendix contents}
\pdfbookmark[0]{Appendix contents}{appendix-contents}
\noindent\begin{tabular*}{\linewidth}{@{}p{.08\linewidth}p{.82\linewidth}@{\extracolsep{\fill}}r@{}}
\toprule
Section & Contents & Page\\\midrule
\hyperref[compact:evaluation]{\ref*{compact:evaluation}} & \hyperref[compact:evaluation]{Scoring and evaluation} & \pageref{compact:evaluation}\\
\hyperref[compact:training]{\ref*{compact:training}} & \hyperref[compact:training]{Data and offline training} & \pageref{compact:training}\\
\hyperref[compact:agents]{\ref*{compact:agents}} & \hyperref[compact:agents]{Agent and comparison protocols} & \pageref{compact:agents}\\
\hyperref[compact:controls]{\ref*{compact:controls}} & \hyperref[compact:controls]{Learned direction analyses} & \pageref{compact:controls}\\
\hyperref[compact:comparisons]{\ref*{compact:comparisons}} & \hyperref[compact:comparisons]{Additional sequence and search comparisons} & \pageref{compact:comparisons}\\
\hyperref[compact:trace]{\ref*{compact:trace}} & \hyperref[compact:trace]{Illustrative search trace} & \pageref{compact:trace}\\
\bottomrule\end{tabular*}
\medskip

\section{Scoring and evaluation}
\label{compact:evaluation}
\subsection{Predictors and score normalization}
\label{appendix-toc-1}
\label{app:evaluation}
Table~\ref{tab:predictor-overview} summarizes the internal and external predictors, including models based on ESM-2 representations \citep{esm22023}. Internal scores guide intervention selection, while external predictors assess the generated candidates after search. 

\begin{table}[H]
\centering\small

\caption{\textbf{Predictor overview.} Released models remain frozen during search; locally trained evaluators are fitted offline.}
\label{tab:predictor-overview}
\begin{tabularx}{\linewidth}{>{\raggedright\arraybackslash}p{.125\linewidth}>{\raggedright\arraybackslash}p{.245\linewidth}>{\raggedright\arraybackslash}X}
\toprule
\multicolumn{1}{l}{\tablehead{Target}} & \multicolumn{1}{c}{\tablehead{Predictor}} & \multicolumn{1}{c}{\tablehead{Training data and method}}\\
\midrule
\multicolumn{3}{l}{\tablegroup{Internal feedback}}\\
\textbf{Activity} & LLAMP~\citep{llamp2025} & Released MIC-trained model combining peptide ESM-2 representations with bacterial genomic features.\\
\addlinespace[2pt]
\textbf{Hemolysis} & Calibrated ConsAMPHemo~\citep{consamphemo2025} & Released HC50 regression model; we fit a gradient-boosted hemolysis classifier using its score and physicochemical features.\\
\addlinespace[2pt]
\textbf{Toxicity} & ToxinPred3~\citep{toxinpred32024} & Released toxic/non-toxic sequence classifier; the native hybrid score supplies toxicity feedback.\\
\addlinespace[3pt]\multicolumn{3}{l}{\tablegroup{External evaluation}}\\
\addlinespace[2pt]
\textbf{Activity} & MIC regression model & Augmented experimental MIC records from four bacterial species; ESM-2 and sequence features feed a neural regressor. Evaluated for \textit{E. coli}.\\
\addlinespace[2pt]
\textbf{Hemolysis} & HemoDL~\citep{hemodl2024} & Double-LightGBM classifier trained on labeled peptide hemolysis data using sequence-derived features and protein language-model embeddings.\\
\addlinespace[2pt]
\textbf{Toxicity} & ESM-2 toxicity classifier & Cleaned, balanced toxicity labels (5,882 training; 1,261 validation; 1,261 test records); logistic regression on ESM-2 features.\\
\bottomrule
\end{tabularx}
\end{table}

\rwh{Normalization.}
We map predictions to utilities in $[0,1]$, with larger values indicating greater activity, non-hemolysis, and non-toxicity. With $\sigma(z)=(1+e^{-z})^{-1}$, activity and external non-hemolysis scores are
\[
u_{\rm act}=\sigma\!\left(\frac{\widehat{\rm pMIC}-5}{0.5}\right),\qquad
u_{\rm nh}^{\rm ext}=\sigma\!\left(\frac{\log_{10}\widehat{\rm HC50}-2}{0.5}\right),
\]
where HC50 is in $\mu$M, capped at 500. Internal non-hemolysis uses the calibrated non-hemolytic probability. Non-toxicity uses $u_{\rm nt}^{\rm int}=\operatorname{clip}(1-t_{\rm hybrid},0,1)$ for ToxinPred3 hybrid score $t_{\rm hybrid}$, and $u_{\rm nt}^{\rm ext}=1-\sigma(-3.35259+6.70517t)$ for ESM-2 toxicity probability $t$.

\rwh{Distribution penalty and screening.}
A Gaussian KDE (bandwidth 0.5) fits 27,733 AMP reference points in standardized GRAVY, Eisenberg hydrophobic moment \citep{eisenberg1982}, charge density $(n_K+n_R-n_D-n_E)/L$, and length. For log density $\ell(x)$ and its tenth-percentile reference threshold $\tau_{10}$,
\[
p(x)=0.3\,\operatorname{clip}\!\left(\tau_{10}-\ell(x),0,1\right),
\qquad \tau_{10}=-6.46797246.
\]
Subtracting this physicochemical support correction from each utility, screening requires all three $u_k(x)-p(x)>0.5$. Primary metrics use unpenalized utilities; penalized sensitivity analysis uses the same screened candidates. Appendix~\ref{app:statistical-protocol} defines frontier metrics and run-wise aggregation.

\subsection{Metrics, aggregation, and resources}
\label{appendix-toc-20}\label{app:statistical-protocol}\label{appendix-toc-9}\label{app:pareto}
\rwh{Budgets.}
Search spans 15--35 residues and ten rounds of 256 candidates. Main comparisons use six independent runs; the two ablation panels use three matched runs each, drawn as disjoint subsets. Bundled ablations and matrix replacement were paired with EASER; targeted input ablations used a separate paired set. Policies share the frozen AMP-adapted DPLM and learned directions. Each AMPGPT2 run retains 2,560 canonical, length-valid, within-run unique candidates, without property preselection (temperature 1, top-$k$ 50, top-$p$ 1, repetition penalty 1.2, maximum 40 new tokens). Duplicates can require additional draws.
\rwh{Frontier metrics.}
For a feasible score set $S$ and reference $r=(0.5,0.5,0.5)$,
\begin{equation}
\operatorname{HV}(S;r)=\operatorname{vol}\left(\bigcup_{s\in S}\prod_{j=1}^3[r_j,s_j]\right).
\end{equation}
We compute exact three-dimensional volume. For the shared empirical reference set $R$,
\begin{equation}
\operatorname{IGD}^{+}(S,R)=\frac{1}{|R|}\sum_{r\in R}\min_{s\in S}\|\max(r-s,0)\|_2.
\end{equation}
The empirical reference is the nondominated union of screened candidates from all ten methods in Tables~\ref{tab:external-generators} and~\ref{tab:main}. It contains 80 points in unpenalized coordinates and is fixed across all main and ablation comparisons. Penalty sensitivity uses a 79-point reference constructed from the same screened candidates in KDE-adjusted coordinates.

\begin{table}[H]
\centering\small
\caption{\textbf{Sensitivity to score penalties.} The search-policy samples match Table~\ref{tab:main}; AMPGPT2 samples match Table~\ref{tab:external-generators}. Only the evaluation coordinates change. Mean $\pm$ sample SD.}
\label{tab:penalized}
\begin{tabularx}{\linewidth}{>{\raggedright\arraybackslash}XS[table-format=1.4(4),table-column-width=0.28\linewidth]S[table-format=1.4(4),table-column-width=0.28\linewidth]}
\toprule
\multicolumn{1}{l}{\tablehead{Method}} & \multicolumn{1}{c}{\tablehead{Exact HV $\uparrow$}} & \multicolumn{1}{c}{\tablehead{IGD+ $\downarrow$}}\\
\midrule
\multicolumn{3}{l}{\tablegroup{Search-policy baselines}}\\
Fixed coefficients & 0.0277 +- 0.0079 & 0.0801 +- 0.0339\\
Random actions & 0.0373 +- 0.0086 & 0.0485 +- 0.0222\\
LinUCB & 0.0373 +- 0.0087 & 0.0544 +- 0.0261\\
NSGA-II & 0.0344 +- 0.0069 & 0.0591 +- 0.0246\\
\addlinespace[3pt]
\multicolumn{3}{l}{\tablegroup{External generators}}\\
AMPGPT2 & 0.0328 +- 0.0027 & 0.0613 +- 0.0096\\
\addlinespace[3pt]
\multicolumn{3}{l}{\tablegroup{Proposed framework}}\\
\rowcolor{tableShade}
\textbf{EASER} & 0.0448 +- 0.0085 & 0.0287 +- 0.0180\\
\bottomrule
\end{tabularx}
\end{table}

\begin{figure}[!htbp]
\centering
\includegraphics[width=\linewidth]{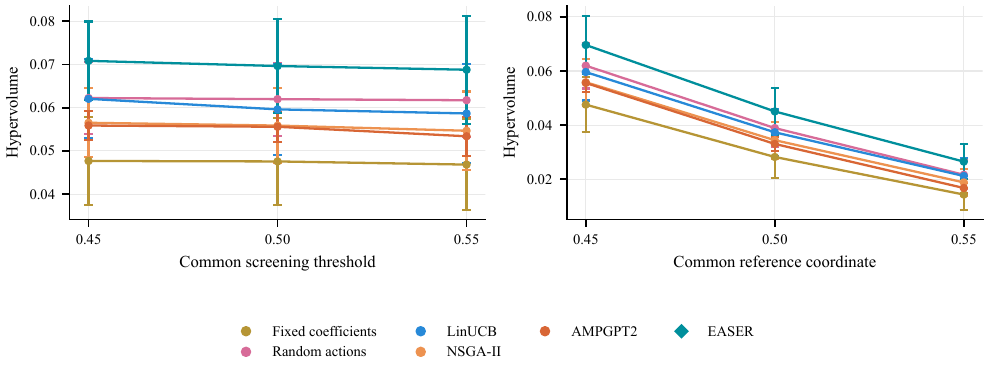}
\caption{\textbf{Screening and reference sensitivity.} Left: vary the common screening threshold with the HV reference fixed at 0.45. Right: vary the common HV reference coordinate with screening fixed at 0.50. All HV calculations use normalized scores before penalty and exact volume. Points and error bars are run means and sample SD.}
\label{fig:sensitivity}
\end{figure}

\rwh{Aggregation and sensitivity.}
Top-$k$ scores are selected independently per property within each run by stable ordering, then averaged equally across runs with sample SD. Source data report $k\in\{10,30,50,100\}$, with the available count specified when fewer than $k$ candidates pass. Nondominated sets are computed per run before projection. Screening thresholds 0.45/0.50/0.55 use fixed HV reference 0.45; reference sensitivity uses the same coordinates at screen 0.50 (Figure~\ref{fig:sensitivity}; mean and sample SD).

\rwh{Resource use.}
Across instrumented runs, 90 LLM calls consumed 1,449,915 input tokens and 472,363 output tokens, averaging about 0.64 million tokens per run. The combined RTX 4090 D usage window was about 0.60 device-hours, including waiting and scoring; concurrent use was counted once. The controller used service-managed API aliases without pinned checkpoint versions.

\section{Data and offline training}
\label{compact:training}
\subsection{Data sources}
\label{appendix-toc-18}
\label{app:data-sources}
Table~\ref{tab:data-sources} summarizes the data sources and principal processing steps. Preference construction and predictor training are described in Appendices~\ref{app:training} and~\ref{app:evaluation}, respectively.

\begin{table}[H]
\centering\small

\caption{\textbf{Data sources and processing.}}
\label{tab:data-sources}
\begin{tabularx}{\linewidth}{>{\raggedright\arraybackslash}p{.205\linewidth}>{\raggedright\arraybackslash}X}
\toprule
\multicolumn{1}{l}{\tablehead{Use}} & \multicolumn{1}{c}{\tablehead{Source and processing}}\\
\midrule
\textbf{Generator adaptation} & Reviewed \href{https://www.uniprot.org/help/uniprotkb}{UniProtKB/Swiss-Prot}~\citep{uniprot2025} short proteins and mature peptides, followed by a curated AMP collection. Sequences are deduplicated and restricted to 7--50 residues; the Swiss-Prot pool is split by sequence clusters.\\
\addlinespace[2pt]
\textbf{Activity preferences} & Experimental MIC records from \href{https://huggingface.co/datasets/ZihengZhou06/AMPBench-MT}{AMPBench-MT}~\citep{ampbench2026}, including DBAASP, GRAMPA, CAMPR4, and DRAMP sources \citep{dbaasp2021,grampa2019,campr42023,dramp2022}. The cleaned \textit{E. coli} subset supplies physicochemically matched preference pairs.\\
\addlinespace[2pt]
\textbf{Safety preferences} & A combined experimental HC50 collection and a curated binary toxicity collection. Sequences are cleaned, filtered, and paired by the corresponding safety labels and physicochemical features.\\
\addlinespace[2pt]
\textbf{Activity evaluator} & Integrated experimental MIC data, supplemented with GRAMPA \citep{grampa2019}, AMP-Diffusion data \citep{ampdiffusion2025}, and published in-vitro measurements. Sequence, species, and MIC records are consolidated for regression.\\
\addlinespace[2pt]
\textbf{AMP reference} & AMP-Diffusion \citep{ampdiffusion2025}, AMPBench-MT \citep{ampbench2026}, and curated peptide activity collections are merged and deduplicated. The 27,733 sequences of length 15--35 provide the physicochemical KDE reference.\\
\addlinespace[2pt]
\textbf{Agent knowledge} & Curated property evidence, amino-acid guidance, motif-risk cards, and intervention strategies are organized into retrievable entries.\\
\bottomrule
\end{tabularx}
\end{table}

\rwh{Code and data availability.} Code, learned property matrices, preprocessing and evaluation scripts, and generated candidates will be released with source identifiers, split assignments, sequences, run configurations, scores, and screening outcomes.

\subsection{Training configuration}
\label{appendix-toc-2}
\rwh{Offline adaptation of the generative prior.}
\label{appendix-toc-3}
\label{app:backbone-adaptation}
DPLM-650M undergoes two offline LoRA stages. Stage 1 trains query and value projections on general short peptides; its step-1800 checkpoint initializes AMP-oriented Stage 2. The selected Stage-2 adapter is merged into the prior, then frozen for output-direction training and search.

Both stages minimize masked denoising cross-entropy with DPLM's corruption mask and token weights:
\begin{equation}
\mathcal L_{\rm adapt}=-\frac{\sum_i m_iw_i\log p_\theta(x_i\mid\widetilde x)}{\max(1,\sum_i m_i)}.
\end{equation}
Here $m_i$ marks prediction positions and $w_i$ is the DPLM loss weight. Both stages use AdamW, batch size 8, four-step accumulation, weight decay 0.01, and gradient clipping at 1.0. Training spans 7--50 residues; online design spans 15--35. Optimizer and sampling configurations are specified in the code.

\begin{table}[H]
\centering\small
\caption{\textbf{Offline backbone adaptation.} Training settings for the two adaptation stages. Counts denote sequence records.}
\begin{tabularx}{\linewidth}{>{\raggedright\arraybackslash}X>{\centering\arraybackslash}p{.26\linewidth}>{\centering\arraybackslash}p{.26\linewidth}}
\toprule
\multicolumn{1}{l}{\tablehead{Setting}} & \multicolumn{1}{c}{\tablehead{\shortstack{Stage 1\\Short peptides}}} & \multicolumn{1}{c}{\tablehead{\shortstack{Stage 2\\AMP-oriented}}}\\
\midrule
Train / validation records & 7,453 / 848 & 8,725 / 1,255\\
LoRA rank / scaling alpha & 8 / 16 & 4 / 8\\
LoRA dropout & 0.05 & 0.10\\
Target projections & Query, value & Query, value\\
Learning rate & $10^{-5}$ & $5\times10^{-6}$\\
Warmup steps & 300 & 150\\
Maximum steps & 3,000 & 1,500\\
Evaluation interval / patience & 300 / 5 & 100 / 3\\
\bottomrule
\end{tabularx}
\end{table}
Validation loss selects the Stage-2 checkpoint; all direction-training and controller experiments share the resulting prior.

\rwh{Output-direction training.}
\label{appendix-toc-4}
\phantomsection\label{app:training}

\rwh{Masked preference objective.}\phantomsection\label{app:preference-objective}
Directions are learned offline from preference pairs matched by coarse physicochemical descriptors. Activity uses minimum inhibitory concentration (MIC), the low-hemolysis direction uses the concentration causing 50\% hemolysis (HC50), and toxicity uses curated source labels.

For each training pair, sample $t$ uniformly from $\{1,\ldots,T\}$ and mask each position eligible in both sequences with probability $t/T$. The pair and both projections share this mask $M$. For the resulting corrupted peptide $x_t$, define the masked conditional score
\begin{equation}
q_W(x;t,M)=\sum_{i\in M}\log p_W(x_i\mid x_t,t).
\label{eq:maskedscore}
\end{equation}
Pairwise training increases the score margin of the preferred member using a logistic preference loss, together with replay and factor regularization. The reference-relative score measures the change in masked conditional log likelihood under the intervention:
\[
s_k(x)=q_{W_0+\Delta W_k}(x)-q_{W_0}(x).
\]
Subscripts identify the output-projection weights used to evaluate $q$; shared corruption arguments are implicit. Training minimizes
\begin{equation}
\mathcal L_k=-\mathbb E\log\sigma\!\left(s_k(x^+)-s_k(x^-)\right)
+\lambda_{\mathrm{rep}}\mathcal L_{\mathrm{replay}}
+\lambda_{\mathrm{norm}}\left(\|U_k\|_F^2+\|V_k\|_F^2\right)+\lambda_{\mathrm{bad}}\mathcal L_{\mathrm{bad}}.
\label{eq:training}
\end{equation}
The expectation averages over training pairs and this corruption distribution. Validation pair accuracy selects the checkpoint before offline norm rescaling. Replay uses denoising negative log likelihood. For the low-hemolysis direction, $\mathcal L_{\mathrm{bad}}$ penalizes likelihood gains on replay sequences from a low-support region of the low-hemolysis physicochemical reference.

\rwh{Inference calibration.} The controller emits $c_k\in[-1,1]$ and shared strength $\eta$, clipped to $\bar\eta=\operatorname{clip}(\eta,0,0.5)$. Matrices are rescaled offline to Frobenius norm approximately 0.225305. A fixed scale factor $g=8$ maps controller outputs to multipliers $\alpha_k=g\bar\eta c_k\in[-4,4]$ in Equation~\ref{eq:intervention}.

All three matrices use the reference-relative masked preference loss (Eq.~\ref{eq:training}). Only $33\times16$ and $1280\times16$ factors are trainable: one starts at zero, the other at Gaussian noise (standard deviation 0.02). AdamW uses learning rate $3\times10^{-4}$, cosine annealing to 5\% of that rate, factor-norm coefficient $10^{-4}$, and gradient clipping at 1.0.

Activity preferences pair \textit{E. coli} records with $5\leq\mathrm{pMIC}<6$ (preferred) and $4\leq\mathrm{pMIC}<5$ (dispreferred), with gaps $\geq$0.8. Both require length 20--35, charge density 0.10--0.45, hydrophobic fraction 0.25--0.55 (AILVFYW), and GRAVY $[-1.2,0.1]$. Supplied identity30 clusters are capped at four preferred and eight dispreferred sequences.

Activity pairs use coarse physicochemical matching and connected-component splits over role-tagged cluster identifiers. Toxicity pairs retain source training, validation, and external-test partitions, pairing non-toxic label 0 with toxic label 1 within the same composition window at lengths 15--35. Low-hemolysis training uses HC50 preferences and replay-based distribution regularization.

\begin{table}[H]
\centering\small
\caption{\textbf{Property-matrix training data and configuration.} Preference-pair counts, selected checkpoints, and trainable factor parameters for each property.}
\label{tab:training}
\begin{tabularx}{\linewidth}{>{\raggedright\arraybackslash}XS[table-format=3,table-column-width=0.095\linewidth]S[table-format=2,table-column-width=0.095\linewidth]S[table-format=2,table-column-width=0.095\linewidth]S[table-format=3,table-column-width=0.18\linewidth]S[table-format=5,table-column-width=0.17\linewidth]}
\toprule
\multicolumn{1}{l}{\tablehead{Property}} & \multicolumn{3}{c}{\tablehead{Preference pairs}} & \multicolumn{1}{c}{\tablehead{\shortstack{Selected\\step}}} & \multicolumn{1}{c}{\tablehead{Parameters}}\\
\cmidrule(lr){2-4}
 & \multicolumn{1}{c}{\tablehead{Train}} & \multicolumn{1}{c}{\tablehead{Valid.}} & \multicolumn{1}{c}{\tablehead{Test}} & & \\
\midrule
Activity & 324 & 14 & 20 & 140 & 21008\\
Low-hemolysis & 406 & 49 & 44 & 170 & 21008\\
Low-toxicity & 200 & 58 & 30 & 210 & 21008\\
\bottomrule
\end{tabularx}
\end{table}
Activity and low-toxicity training use batch size 8, replay weight 0.1, and at most 400 steps; low-hemolysis uses 16, 0.25, and 500. Generated property responses and random-matrix controls assess effectiveness (Table~\ref{tab:cold-directions}; Appendix~\ref{app:full-position-grid}).

Low-hemolysis replay combines high-HC50 sequences with sequences from a low-support region of its physicochemical reference. Its additional loss is $0.6\,\operatorname{softplus}(5g_B)$, where $g_B$ is the batch-level denoising NLL decrease from base to intervened projection. Base and adapted evaluations use independently sampled corruptions, so the replay penalty regularizes their denoising likelihood difference.

We also express the effective strength as $\beta_k=\bar\eta c_k$, with the calibrated projection multiplier given by $\alpha_k=8\beta_k$.

\section{Agent and comparison protocols}
\label{compact:agents}
\subsection{Evidence and memory}
\label{appendix-toc-5}
\rwh{Knowledge graph and retrieval architecture.}
\label{appendix-toc-6}
\label{app:knowledge-architecture}
The evidence resource contains 12,445 property-evidence records and 75 residue/motif guidance records. Deduplicating source identifiers and identical files leaves 3,582 source records. Each entry links property, organism, residue, or motif statements to their sources and confidence assessments. Retrieval combines lexical relevance with sequence and physicochemical context to identify supporting and adverse evidence for each property. Literature evidence and within-run experience are maintained separately.

\rwh{Agent roles and episodic context.}
\label{appendix-toc-7}
\label{app:agents}
\begin{algorithm}[H]
\caption{EASER: online Probe-and-Steer}
\label{alg:online}
\small
\begin{algorithmic}[1]
\Require Frozen generator $G$ and calibrated directions $\{\Delta W_k\}$;
\Statex \hspace{\algorithmicindent} internal evaluator $F$, evidence resource $K$, initial sequence records $R$
\State Initialize online archive $A$ and episodic context $H$ from available records
\For{$r=1,\ldots,10$}
  \State $X_{\rm novo}\gets\Call{DeNovoSample}{G,32}$; evaluate $F(X_{\rm novo})$
  \State $S\gets\Call{Build\_Context}{R,A,H,X_{\rm novo}}$ \Comment{Sequences, properties, and experience}
  \State $E_k\gets\Call{AggregateEvidence}{k,S,K}$ for $k\in\{a,h,t\}$
  \State $\{a_j\}_{j=1}^{4}\gets\Call{ControllerPropose}{S,\{E_k\}}$ \Comment{$a_j=(P_j,M_j,\bm c_j,\eta_j)$}
  \State $X_j^{\rm probe}\gets\Call{Generate}{G,a_j,16,\{\Delta W_k\}}$ for each $j$
  \State $Y^{\rm probe}\gets F(\{X_j^{\rm probe}\}_{j=1}^{4})$
  \State $\bm n\gets\Call{ControllerSteer}{S,\{a_j\},Y^{\rm probe}}$, with $n_j\geq0$, $\sum_jn_j=160$
  \State $X_j^{\rm steer}\gets\Call{Generate}{G,a_j,n_j,\{\Delta W_k\}}$ for each $j$
  \State $X\gets X_{\rm novo}\cup\bigcup_j(X_j^{\rm probe}\cup X_j^{\rm steer})$; score with $F$
  \State Update $R,A$ with outcomes, retaining eligible candidates for further refinement
  \State Record actions and outcomes; reflect and update $H$ under the memory policy
\EndFor
\State \Return All generated sequences and execution records for post-hoc external evaluation
\end{algorithmic}
\end{algorithm}
\rwh{Evidence aggregators.}
Three evidence aggregators deterministically synthesize supporting, adverse, and mixed evidence into property-specific recommendations, confidence, and risks; conflicts reduce confidence. The controller makes the joint design decision.

\rwh{Main controller.}
The DeepSeek Flash controller (service alias \texttt{deepseek-flash}) proposes four interventions per round from property evidence, visible sequences, and prior outcomes. Probe feedback guides further sampling among them.

\rwh{History and reflection.}
After each round, the controller summarizes successes and failures, suggesting parent, mask, and coefficient changes. Summaries and numerical outcomes enter the next proposal context. Memory is initialized independently for each run and accumulates within-run experience. Appendix~\ref{app:search-case} traces reflection into subsequent proposals.

\rwh{Execution and budgets.}
Algorithm~\ref{alg:online} allocates 32/64/160 samples per round. The frozen 650M backbone uses 63,024 learned factor parameters stored as 126,720 matrix entries, without online gradients. Candidates failing final screening remain eligible parents.

\subsection{Numerical controllers}
\label{appendix-toc-8}
\label{app:controllers}
Numerical controllers share EASER's generator, matrices, and length range. A run comprises ten rounds of 256 candidates: 32 de novo, 64 Probe, and 160 Steer per round.

\rwh{Shared action space.}
Both baselines select from all candidates in the same design context. Random-key encoding supports any parent subset, including none for de novo sampling, and any nonempty position subset up to half each parent's length. Coefficients span $[-1,1]^3$; strength spans $[0,0.5]$. The prior favors small parent sets with nonzero support for every legal subset. 

\rwh{NSGA-II.}
Online NSGA-II uses nondominated sorting and crowding \citep{nsga22002}. Round one probes four random actions; later rounds probe two survivors, remapped to the current context and reevaluated, plus two offspring. Simulated binary crossover uses index 15; polynomial mutation uses index 20 and probability $1/d$ for encoding dimension $d$. Twenty percent of offspring proposals are fresh random actions; each round selects two survivors.

\rwh{LinUCB.} The contextual-bandit rule of \citet{linucb2010} selects interventions.
Each round pools 256 fresh actions with retained and mutated survivors, without generating peptides. Features summarize effective coefficients, parent properties and composition, masked residues, position statistics, and context. An identity ridge prior and exploration multiplier 1 rank actions by reward plus uncertainty; four distinct actions are probed. Reward is $0.4a+0.3h+0.3t$. Distinct sequence-level actions can share the same feature vector.

\rwh{Feedback and allocation.}
Both algorithms batch-center KDE-penalized internal activity, non-hemolysis, and non-toxicity objectives $(a,h,t)$, without KL penalties or external evaluation during search. Steer allocation follows nondomination ranks in eight-sample quanta, with at least eight and at most 120 per proposal, conserving the 160-sample budget. 

\subsection{Matched-information comparison}
\label{appendix-toc-23}\label{app:grounded-reasoning}
The comparison in Table~\ref{tab:grounded-reasoning} uses the ten pre-Probe states from each campaign. Both policies receive identical sequences, internal scores, aggregator claims, and history, augmented with residue counts, indexed sequences, and descriptors for this comparison. Explicit formulation states four hypotheses with supporting observations and risks; direct generation returns four actions. Both use the same DeepSeek Flash controller at temperature zero with the API thinking option disabled. The generator and matrices are fixed, and sampling schedules are paired. Each action produces 16 candidates, giving 64 candidates per state and 1,920 per policy.

The hypotheses cited 360 observations, all consistent with the supplied values. After bounded correction, one proposal had a mismatch between its mask and rationale. Full masking specifies de novo sampling. We measure each state's HV gain over all pre-Probe candidates, including the current round's initial de novo batch. Both policies use equal sampling allocations, the common external screen, and the unpenalized HV reference $(0.5,0.5,0.5)$.

\subsection{Input and memory ablations}
\label{appendix-toc-25}
\label{app:targeted-controls}
The targeted ablations in Table~\ref{tab:targeted-feedback} retain the generator, property matrices, anchor-selection procedure, evidence retrieval, and Probe-and-Steer. The reflection-feedback ablation clears reflection fields from LLM inputs. The round-history-details ablation removes action and outcome fields from history entries, retaining round identifiers, reflection texts, retrieved episodes, promoted rules, and historical anchors.

The memory-and-anchor ablation in Table~\ref{tab:ablation} removes round-history inputs, experience storage and retrieval, rule promotion, and historical anchor reuse. The anchor pool consists of the current round's 32 de novo candidates. Reflection summaries are generated each round and excluded from subsequent proposal contexts. The archive collects candidates for final evaluation. Each panel pairs its ablations with EASER.

\subsection{Random-matrix replacement}
\label{appendix-toc-15}\label{app:full-search-random}
Each $\Delta W_k=U_k\Sigma_kV_k^\top$ is replaced by $Q_k\Sigma_kR_k^\top$ with randomized orthonormal factors, preserving norm and spectrum. Fixed random matrix sets are paired with matched runs; generator, agent, feedback, screen, reference front, and 2,560-candidate budget are unchanged, while actions adapt to feedback. Learned matrices achieved higher mean HV and lower mean IGD+ under both score conventions (Table~\ref{tab:ablation}). Random replacement performed better in one of the matched runs and produced more screened candidates overall (1,076 versus 871).

\section{Learned direction analyses}
\label{compact:controls}
\subsection{Residue preferences and context}
\label{appendix-toc-10}\label{app:matrix}
We compared learned and spectrum-matched random matrices on identical zero-coefficient hidden states from 500 peptides at 25\%, 50\%, 75\%, and 100\% masking. Two 50\%-mask controls shuffled residue order; the stricter shuffle preserved total and visible composition, length, and masked positions.

For each masked position, let $z_n$ be $3.2\Delta W_kh$ over the 20 amino-acid logits after removing its across-residue mean. The fraction $N\|\bar z\|_2^2/\sum_n\|z_n\|_2^2$ measures squared shift energy captured by the mean vector, where $N$ counts positions and $\bar z$ is their mean. At 50\% masking (6,258 positions), these fractions were 96.4\%, 60.4\%, and 96.9\% for activity, non-hemolysis, and non-toxicity. F, K, and V had the largest increases at 99.6\%, 98.7\%, and 100.0\% of positions. Mean cosine similarities after the stricter shuffle were 0.9973, 0.9781, and 0.9994. Context changed shift magnitude more than direction, especially for the activity and low-toxicity directions.

\begin{figure}[!htbp]
\centering
\includegraphics[width=0.92\linewidth]{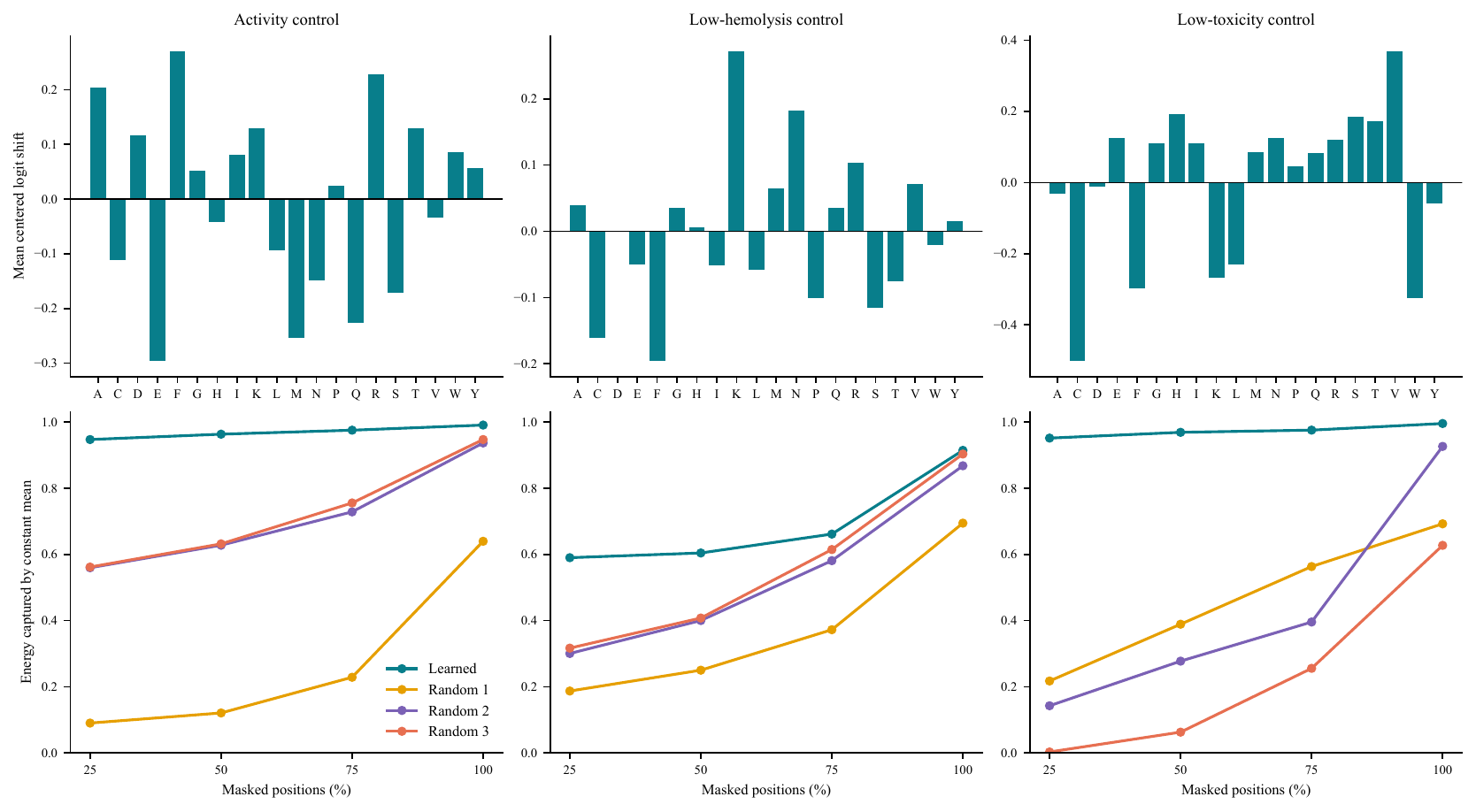}
\caption{\textbf{Residue preferences and contextual variation of output interventions.} Top: mean centered logit shifts for learned matrices at 50\% masking across 6,258 masked positions from 500 peptides. Bottom: fraction of squared shift energy captured by a constant mean vector, evaluated separately at each mask fraction. Each random set preserves the corresponding learned matrix's spectrum. }
\label{fig:matrix-context}
\end{figure}

\subsection{Single-matrix dose responses}
\label{appendix-toc-14}\label{app:component-diagnostics}
Single-matrix responses are measured by de novo sampling from fully masked sequences. Learned matrices and three spectrum-matched random sets use the same 500 lengths and paired sampling schedules, including identical zero-coefficient sequences. Each scan uses $c\in\{-1,-0.5,0,0.25,0.5,0.75,1\}$ and $\eta=0.4$ ($\alpha=3.2c$), with the other coefficients zero. Sharing the zero-coefficient condition gives 19 conditions per set and 38,000 generated records overall. Means are computed before screening and retain duplicate draws.

\begin{table}[H]
\centering\small
\sisetup{round-mode=places,round-precision=4}
\caption{\textbf{Learned target means across the full signed dose grid.} Normalized scores before penalty, 500 draws per condition; only the named property matrix is active.}
\label{tab:cold-dose}
\begin{tabular*}{\linewidth}{@{\extracolsep{\fill}}S[table-format=+1.2,table-column-width=0.12\linewidth,round-mode=none]S[table-format=1.5,table-column-width=0.255\linewidth]S[table-format=1.5,table-column-width=0.255\linewidth]S[table-format=1.5,table-column-width=0.255\linewidth]}
\toprule
\multicolumn{1}{c}{\tablehead{$c$}} & \multicolumn{1}{c}{\tablehead{\shortstack{Activity\\score}}} & \multicolumn{1}{c}{\tablehead{\shortstack{Non-hemolysis\\score}}} & \multicolumn{1}{c}{\tablehead{\shortstack{Non-toxicity\\score}}}\\
\midrule
-1 & 0.24837 & 0.44792 & 0.66840\\
-0.5 & 0.26753 & 0.44699 & 0.71803\\
0 & 0.25784 & 0.45066 & 0.75425\\
0.25 & 0.25859 & 0.45472 & 0.77322\\
0.5 & 0.28017 & 0.46308 & 0.79254\\
0.75 & 0.28755 & 0.47216 & 0.79798\\
1 & 0.28972 & 0.48236 & 0.82420\\
\bottomrule
\end{tabular*}
\end{table}

\begin{table}[H]
\centering\small
\sisetup{round-mode=places,round-precision=4}
\caption{\textbf{Cross-property effects at $c=1$.} Learned $c=1$ minus the zero-coefficient setting, averaged over all 500 de novo draws before screening.}
\label{tab:cold-cross}
\begin{tabularx}{\linewidth}{>{\raggedright\arraybackslash}XS[table-format=+1.5,table-column-width=0.24\linewidth]S[table-format=+1.5,table-column-width=0.24\linewidth]S[table-format=+1.5,table-column-width=0.24\linewidth]}
\toprule
\multicolumn{1}{l}{\tablehead{Active matrix}} & \multicolumn{1}{c}{\tablehead{\shortstack{Activity\\score}}} & \multicolumn{1}{c}{\tablehead{\shortstack{Non-hemolysis\\score}}} & \multicolumn{1}{c}{\tablehead{\shortstack{Non-toxicity\\score}}}\\
\midrule
Activity & +0.03188 & -0.00133 & +0.02242\\
Low-hemolysis & +0.02132 & +0.03170 & +0.04748\\
Low-toxicity & -0.01280 & +0.02733 & +0.06996\\
\bottomrule
\end{tabularx}
\end{table}
Positive activity and low-hemolysis directions increased mean charge density from 0.0459 to 0.1081 and 0.1066; low-toxicity direction reduced it to 0.0305. Target gains remained positive after KDE subtraction.

\subsection{Joint coefficient responses}
\label{appendix-toc-21}\label{app:full-position-grid}

\begin{table}[H]
\centering\small
\sisetup{round-mode=places,round-precision=4}
\caption{\textbf{Target responses under full de novo sampling.} Mean target-score change before penalty from $c=0$ to $c=1$, with the other coefficients zero. Each condition contains 500 draws; the three random sets preserve each learned matrix\textquotesingle s singular-value spectrum and norm.}
\label{tab:cold-directions}
\begin{tabularx}{\linewidth}{>{\raggedright\arraybackslash}XS[table-format=+1.5,table-column-width=0.24\linewidth]S[table-format=+1.5,table-column-width=0.24\linewidth]S[table-format=+1.5,table-column-width=0.24\linewidth]}
\toprule
\multicolumn{1}{l}{\tablehead{Matrix set}} & \multicolumn{1}{c}{\tablehead{\shortstack{Activity\\score}}} & \multicolumn{1}{c}{\tablehead{\shortstack{Non-hemolysis\\score}}} & \multicolumn{1}{c}{\tablehead{\shortstack{Non-toxicity\\score}}}\\
\midrule
\rowcolor{tableShade}
\textbf{Learned} & +0.03188 & +0.03170 & +0.06996\\
Random 1 & +0.00261 & +0.00448 & +0.01503\\
Random 2 & +0.00461 & +0.00701 & +0.00057\\
Random 3 & +0.00577 & -0.00398 & +0.01226\\
\bottomrule
\end{tabularx}
\end{table}
The joint sweep reuses the 500-draw length and sampling schedule with all positions masked and $\eta=0.4$ ($\alpha_k=3.2c_k$). Coefficients $c_k\in\{-1,-0.5,0,0.5,1\}$ yield 125 settings per set and 250,000 records across learned and three spectrum-matched random sets. Means include all draws before screening; zero-coefficient sequences are identical.

\begin{table}[H]
\centering\small
\sisetup{round-mode=places,round-precision=4}
\caption{\textbf{Best objective-specific mean gains in the joint coefficient sweep.} Changes in KDE-adjusted utilities relative to the zero-coefficient setting. Each cell is the maximum over 125 combinations evaluated on the same 500-draw schedule. The three columns can select different combinations.}
\label{tab:full-position-grid}
\begin{tabularx}{\linewidth}{>{\raggedright\arraybackslash}XS[table-format=+1.5,table-column-width=0.24\linewidth]S[table-format=+1.5,table-column-width=0.24\linewidth]S[table-format=+1.5,table-column-width=0.24\linewidth]}
\toprule
\multicolumn{1}{l}{\tablehead{Matrix set}} & \multicolumn{1}{c}{\tablehead{\shortstack{Activity\\score}}} & \multicolumn{1}{c}{\tablehead{\shortstack{Non-hemolysis\\score}}} & \multicolumn{1}{c}{\tablehead{\shortstack{Non-toxicity\\score}}}\\
\midrule
\rowcolor{tableShade}
\textbf{Learned} & +0.06525 & +0.03074 & +0.08807\\
Random 1 & +0.01832 & +0.01226 & +0.01455\\
Random 2 & +0.01296 & +0.01353 & +0.02000\\
Random 3 & +0.01834 & +0.01517 & +0.01579\\
\bottomrule
\end{tabularx}
\end{table}
Learned objective-specific maxima occur at $(1,0,-1)$, $(0,1,1)$, and $(0,0.5,1)$ for activity, non-hemolysis, and non-toxicity, respectively. The all-positive setting $(1,1,1)$ changes adjusted means by $+0.00789$, $+0.01192$, and $+0.07955$, so it maximizes none of the three.

\begin{table}[H]
\centering\small
\caption{\textbf{Response coverage and pooled candidate quality in the full grid.} The first three columns give percentages of 124 nonzero settings whose adjusted target mean increased by more than 0.01. HV pools 62,500 draws per set, screens on adjusted scores, and uses unpenalized normalized coordinates with reference $(0.5,0.5,0.5)$. }
\label{tab:matrix-coverage}
\begin{tabularx}{\linewidth}{>{\raggedright\arraybackslash}XS[table-format=2.1,table-column-width=0.185\linewidth]S[table-format=2.1,table-column-width=0.185\linewidth]S[table-format=2.1,table-column-width=0.185\linewidth]S[table-format=1.5,table-column-width=0.185\linewidth]}
\toprule
\multicolumn{1}{l}{\tablehead{Matrix set}} & \multicolumn{1}{c}{\tablehead{\shortstack{Activity\\score (\%)}}} & \multicolumn{1}{c}{\tablehead{\shortstack{Non-hemolysis\\score (\%)}}} & \multicolumn{1}{c}{\tablehead{\shortstack{Non-toxicity\\score (\%)}}} & \multicolumn{1}{c}{\tablehead{Pooled HV}}\\
\midrule
\rowcolor{tableShade}
\textbf{Learned} & 41.9 & 26.6 & 44.4 & 0.0496\\
Random 1 & 10.5 & 2.4 & 8.9 & 0.0369\\
Random 2 & 4.0 & 3.2 & 12.9 & 0.0352\\
Random 3 & 14.5 & 3.2 & 12.9 & 0.0384\\
\bottomrule
\end{tabularx}
\end{table}
Learned directions produced a wider response range that included adverse trade-offs. Twenty learned settings increased all three adjusted means over the zero-coefficient setting, compared with 30, 29, and 46 settings for the random sets. Among 112 learned multi-matrix settings, 55 response vectors were not weakly dominated by any single-matrix or zero-coefficient vector. Settings using multiple matrices accounted for 14 of the 15 nondominated mean vectors in the full grid.

Single-matrix adjusted maxima were 0.03451, 0.02145, and 0.06407 for activity, non-hemolysis, and non-toxicity, respectively; joint maxima were 0.06525, 0.03074, and 0.08807. The axis subset and full joint grid contained 13 and 125 settings, respectively. Learned and random matrices were compared on the same 125-setting grid. Source data include condition means, draw-level scores, and response thresholds of 0 and 0.005 in addition to 0.01.

\begin{figure}[!htbp]
\centering
\includegraphics[width=\linewidth]{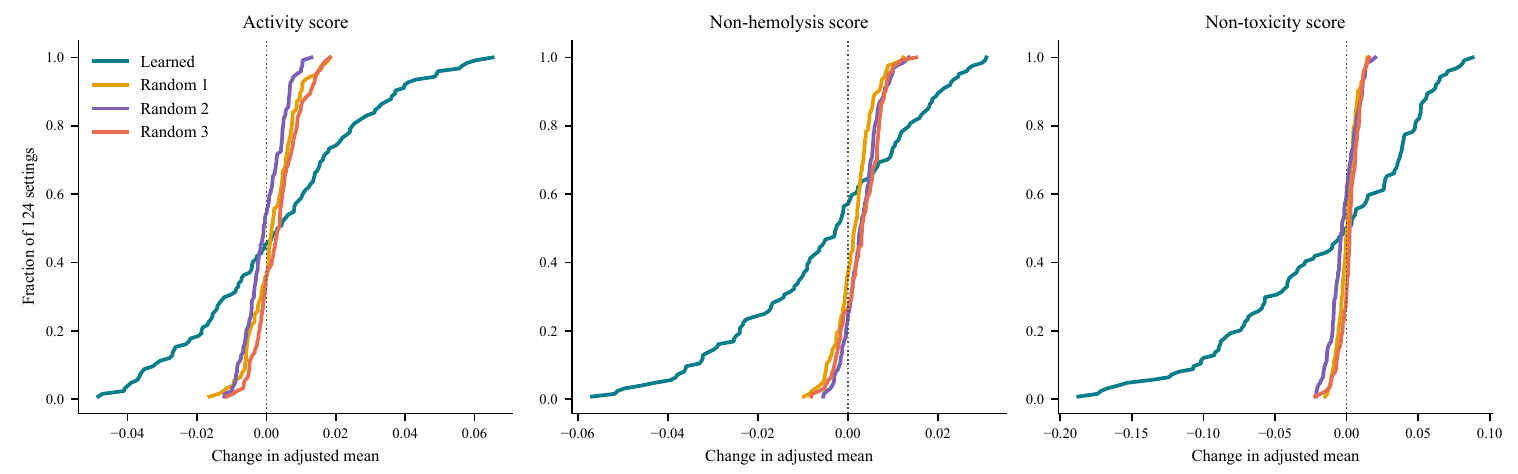}
\caption{\textbf{Distribution of direction responses.} Empirical cumulative distributions of adjusted mean changes from the zero-coefficient setting across 124 nonzero settings, each with 500 draws. Learned directions produce larger beneficial and adverse responses.}
\label{fig:matrix-space}
\end{figure}

\section{Additional sequence and search comparisons}
\label{compact:comparisons}
\subsection{Property-specific selection}
\label{appendix-toc-12}
\label{app:topk}

\begin{table}[H]
\centering\small
\caption{\textbf{Search-policy top ten after screening.} Each column ranks by its own normalized score before penalty within each run. Mean $\pm$ sample SD; every included run had at least ten feasible candidates.}
\label{tab:top10}
\begin{tabularx}{\linewidth}{>{\raggedright\arraybackslash}XS[table-format=1.3(3),table-column-width=0.23\linewidth]S[table-format=1.3(3),table-column-width=0.23\linewidth]S[table-format=1.3(3),table-column-width=0.23\linewidth]}
\toprule
\multicolumn{1}{l}{\tablehead{Method}} & \multicolumn{1}{c}{\tablehead{\shortstack{Activity\\score $\uparrow$}}} & \multicolumn{1}{c}{\tablehead{\shortstack{Non-hemolysis\\score $\uparrow$}}} & \multicolumn{1}{c}{\tablehead{\shortstack{Non-toxicity\\score $\uparrow$}}}\\
\midrule
\multicolumn{4}{l}{\tablegroup{Search-policy baselines}}\\
Fixed coefficients & 0.842 +- 0.089 & 0.667 +- 0.036 & 0.959 +- 0.003\\
Random actions & 0.837 +- 0.026 & 0.717 +- 0.039 & 0.963 +- 0.003\\
LinUCB & 0.861 +- 0.027 & 0.725 +- 0.026 & 0.961 +- 0.002\\
NSGA-II & 0.830 +- 0.060 & 0.694 +- 0.017 & 0.961 +- 0.001\\
\addlinespace[3pt]
\multicolumn{4}{l}{\tablegroup{Proposed framework}}\\
\rowcolor{tableShade}
\textbf{EASER} & 0.901 +- 0.067 & 0.731 +- 0.024 & 0.963 +- 0.003\\
\bottomrule
\end{tabularx}
\end{table}
Source data provide top-$k$ profiles for $k\in\{10,30,50,100\}$, companion-property scores, and the available counts for each selected set.

\subsection{Reference-front coverage}
\label{appendix-toc-16}\label{app:coverage}
Screened sequences are deduplicated within each run per condition, and each run\textquotesingle s three-objective front $F_s$ is computed before projection. With the fixed empirical reference $R$ used for IGD+, coverage is
\begin{equation}
C_\epsilon(F_s;R)=\frac{1}{|R|}\sum_{r\in R}\mathbf{1}\!\left[\min_{z\in F_s}\|r-z\|_\infty\leq\epsilon\right].
\end{equation}
Coverage measures proximity to observed trade-offs. Occupancy counts score-space cubes anchored at 0.5 in each coordinate. The reference includes evaluated methods.

\begin{table}[H]
\centering\small
\caption{\textbf{Within-run Pareto counts and coverage.} Mean $\pm$ sample SD. Coverage uses $\epsilon=0.025$ and is reported in percent; occupied cells have side length 0.05. Sequence counts are deduplicated within each run.}
\label{tab:coverage}

\begin{tabularx}{\linewidth}{>{\raggedright\arraybackslash}XS[table-format=2.1(3),table-column-width=0.23\linewidth]S[table-format=2.1(2),table-column-width=0.23\linewidth]S[table-format=2.1(3),table-column-width=0.23\linewidth]}
\toprule
\multicolumn{1}{l}{\tablehead{Method}} & \multicolumn{1}{c}{\tablehead{\shortstack{Pareto\\sequences}}} & \multicolumn{1}{c}{\tablehead{\shortstack{Occupied\\cells}}} & \multicolumn{1}{c}{\tablehead{Coverage (\%)}}\\
\midrule
\multicolumn{4}{l}{\tablegroup{Search-policy baselines}}\\
Fixed coefficients & 23.3 +- 15.3 & 15.2 +- 7.9 & 11.9 +- 10.8\\
Random actions & 15.5 +- 6.2 & 13.8 +- 3.9 & 16.9 +- 11.0\\
LinUCB & 26.0 +- 12.6 & 17.8 +- 6.0 & 18.3 +- 8.3\\
NSGA-II & 12.7 +- 4.2 & 11.7 +- 3.8 & 11.2 +- 4.5\\
\addlinespace[3pt]
\multicolumn{4}{l}{\tablegroup{External generators}}\\
AMPGPT2 & 20.3 +- 2.6 & 18.2 +- 1.9 & 9.8 +- 4.0\\
\addlinespace[3pt]
\multicolumn{4}{l}{\tablegroup{Proposed framework}}\\
\rowcolor{tableShade}
\textbf{EASER} & 28.7 +- 14.9 & 21.3 +- 8.2 & 36.0 +- 21.9\\
\bottomrule
\end{tabularx}
\end{table}

\subsection{Physicochemical profiles}
\label{appendix-toc-13}\label{app:external-physchem}
Figure~\ref{fig:external-physchem} compares the physicochemical profiles of candidates passing the common screen with those of the real-AMP reference. Descriptors retain their original units; source data include candidate identities, per-run summaries, amino-acid composition, and charge at pH 7.4.

\subsection{Search progress}
\label{appendix-toc-22}\label{app:search-progress}
\begin{figure}[!p]
\centering
\includegraphics[width=0.82\linewidth]{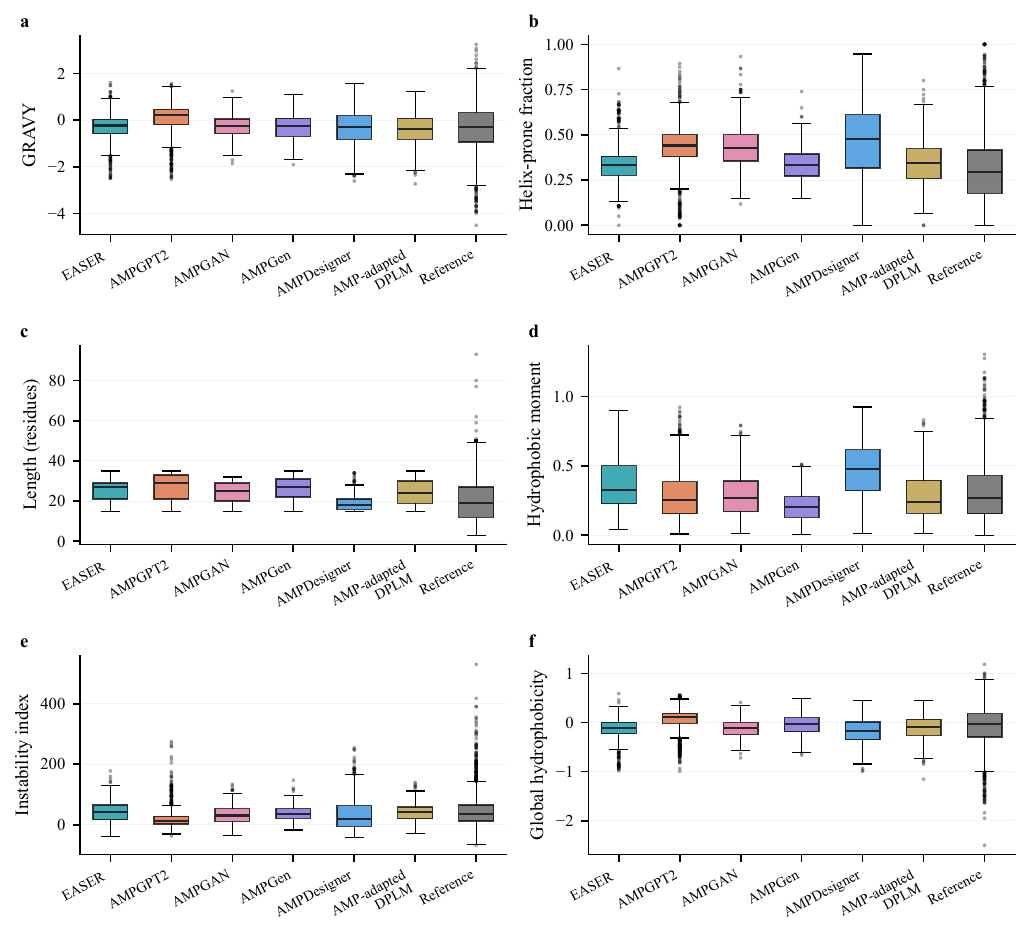}
\caption{\textbf{Physicochemical profiles after screening.} Candidate records were collected per generator before screening; retained counts: EASER 871, AMPGPT2 846, AMPGAN v2 530, AMPGen 189, AMP-Designer 691, and AMP-adapted DPLM 293. The real-AMP reference has 2,000 sequences, separate from the 27,733-point KDE reference. Boxes show median and interquartile range; whiskers extend to 1.5 IQR and points mark remaining observations. Helix-prone fraction uses EMALK; hydrophobic moment uses the Eisenberg scale and $100^\circ$ angle.}
\label{fig:external-physchem}
\vspace{5pt}
\centering
\includegraphics[width=0.55\linewidth]{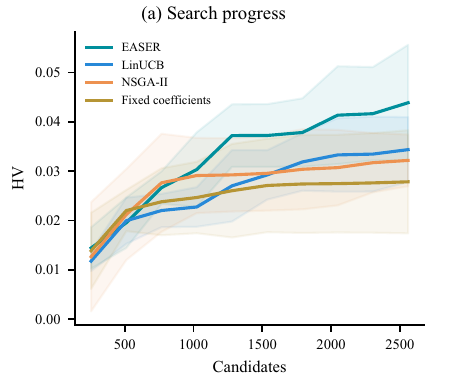}
\caption{\textbf{Search progress.} Cumulative screened HV across the matched diagnostic panel; lines show means and shaded bands denote $\pm$ sample SD. Coordinates are normalized scores before penalty, with the reference from Table~\ref{tab:main}. Endpoint and ablation comparisons appear in Tables~\ref{tab:main} and~\ref{tab:ablation}.}
\label{fig:search-progress}
\end{figure}

Figure~\ref{fig:search-progress} reconstructs cumulative external HV from recorded generation rounds. The same learned matrices support all controllers; numerical policies choose parents, masks, coefficients, and strength. The fixed-coefficient baseline uses predefined $(0.5,0.5,0.5)$ coefficients and an editing policy throughout search. Across the main-search runs (Table~\ref{tab:main}), EASER's final mean HV exceeds LinUCB and NSGA-II by 20.7\% and 30.9\% under the same retained-candidate budget.

\FloatBarrier
\section{Illustrative search trace}
\label{compact:trace}

\label{appendix-toc-17}\label{app:search-case}\label{sec:case}
\rwh{Selection and evidence.}
We examine an offspring from round 3, proposal 3. A fixed ordering by round, proposal, and row selected it from 32 mask-consistent, single-parent offspring on the final externally screened fronts of the diagnostic panel. The deterministic aggregators (Appendix~\ref{app:agents}) each cited six records. Source identifiers appeared in aggregator recommendations and were omitted from the controller's proposal. Recommendations for activity, non-hemolysis, and non-toxicity were $0.350$, $-0.248$, and $-0.233$, with respective confidence scores of $0.423$, $0.689$, and $0.421$. Activity and toxicity evidence conflicted, whereas hemolysis evidence supported the aggregator's recommendation.

\rwh{Hypothesis and execution.}
The parent's internal scores were $(0.697,0.402,1.000)$. The controller targeted its C-terminal hydrophobic/aromatic region with coefficients $(0.65,0.80,0.90)$, hypothesizing reduced hemolysis with retained activity and recording the risk of activity loss. All nine substitutions respected the mask; both sequences have 27 residues:
\begin{center}\small
\begin{tabular}{@{}ll@{}}
Parent & \texttt{MKKKKGTKRKNLRVKPWTPLVIHWHSS}\\
Offspring & \texttt{MKKKKGTKRKNTRGKPVTPAMPVNPSS}\\
\end{tabular}
\end{center}
Each proposal received 16 probes; Steer then allocated $(0,0,40,120)$ samples. Proposal 4 used balanced de novo sampling with fixed coefficients.

\begin{table}[H]
\centering\small
\caption{\textbf{Recorded Probe feedback and subsequent budgets.} Scores are internal feedback means under the online scoring convention; penalties can make values negative. Each proposal used 16 probes.}
\label{tab:search-case-probe}
\begin{tabularx}{\linewidth}{>{\raggedright\arraybackslash}XS[table-format=+1.3,table-column-width=0.205\linewidth]S[table-format=+1.3,table-column-width=0.205\linewidth]S[table-format=1.3,table-column-width=0.205\linewidth]S[table-format=3,table-column-width=0.14\linewidth]}
\toprule
\multicolumn{1}{l}{\tablehead{Proposal}} & \multicolumn{1}{c}{\tablehead{\shortstack{Activity\\score}}} & \multicolumn{1}{c}{\tablehead{\shortstack{Non-hemolysis\\score}}} & \multicolumn{1}{c}{\tablehead{\shortstack{Non-toxicity\\score}}} & \multicolumn{1}{c}{\tablehead{\shortstack{Steer\\samples}}}\\
\midrule
1 & 0.338 & -0.020 & 0.148 & 0\\
2 & -0.036 & 0.642 & 0.283 & 0\\
3 & 0.226 & 0.544 & 0.528 & 40\\
4 & 0.279 & 0.628 & 0.917 & 120\\
\bottomrule
\end{tabularx}
\end{table}

\begin{table}[H]
\centering\small
\caption{\textbf{Parent and offspring under the common external evaluation.} Before penalty denotes normalized scores before KDE subtraction; adjusted scores determine eligibility. The three utility dimensions are all maximized.}
\label{tab:search-case-external}
\begin{tabularx}{\linewidth}{>{\raggedright\arraybackslash}X>{\centering\arraybackslash}p{0.14\linewidth}S[table-format=1.4,table-column-width=0.19\linewidth]S[table-format=1.4,table-column-width=0.19\linewidth]S[table-format=1.4,table-column-width=0.19\linewidth]}
\toprule
\multicolumn{1}{l}{\tablehead{Score convention}} & \multicolumn{1}{c}{\tablehead{Sequence}} & \multicolumn{1}{c}{\tablehead{\shortstack{Activity\\score}}} & \multicolumn{1}{c}{\tablehead{\shortstack{Non-hemolysis\\score}}} & \multicolumn{1}{c}{\tablehead{\shortstack{Non-toxicity\\score}}}\\
\midrule
Before penalty & Parent & 0.6487 & 0.4792 & 0.9575\\
Before penalty & Offspring & 0.6456 & 0.6811 & 0.9559\\
\addlinespace[3pt]\cmidrule(lr){1-5}
After penalty & Parent & 0.6487 & 0.4792 & 0.9575\\
After penalty & Offspring & 0.5368 & 0.5723 & 0.8471\\
\bottomrule
\end{tabularx}
\end{table}
\rwh{Outcome and reflection.}
The offspring passed all external screening thresholds after a KDE penalty of 0.1088 and entered the final external front; its parent failed the non-hemolysis threshold. Its internal adjusted scores were $(0.2524,0.6946,0.6312)$, and the proposal contributed no new online nondominated candidates.

Reflection recommended selecting another parent for the C-terminal intervention and retesting balanced de novo sampling. The next round selected a parent from proposal 4's lineage, citing that branch's earlier frontier expansion.

\end{document}